\documentclass[10pt,twocolumn,letterpaper]{article}
\usepackage{wrapfig}
\usepackage{cvpr}
\usepackage{multirow}
\usepackage[accsupp]{axessibility}
\usepackage[dvipsnames]{xcolor}

\definecolor{cvprblue}{rgb}{0.21,0.49,0.74}
\usepackage[pagebackref,breaklinks,colorlinks,citecolor=cvprblue]{hyperref}

\usepackage{float}

\def\paperID{16937}
\def\confName{CVPR}
\def\confYear{2024}

\usepackage{xcolor}
\title{\textit{SPIN:} Simultaneous Perception, Interaction and Navigation}

\author{Shagun Uppal 
\and
Ananye Agarwal
\and 
Haoyu Xiong
\and
Kenneth Shaw
\and
Deepak Pathak
\and
Carnegie Mellon University
}

\begin{document}

\twocolumn[{%
\renewcommand\twocolumn[1][]{#1}%
\maketitle
\begin{center}
    \centering
    \captionsetup{type=figure}
    \includegraphics[width=\linewidth]{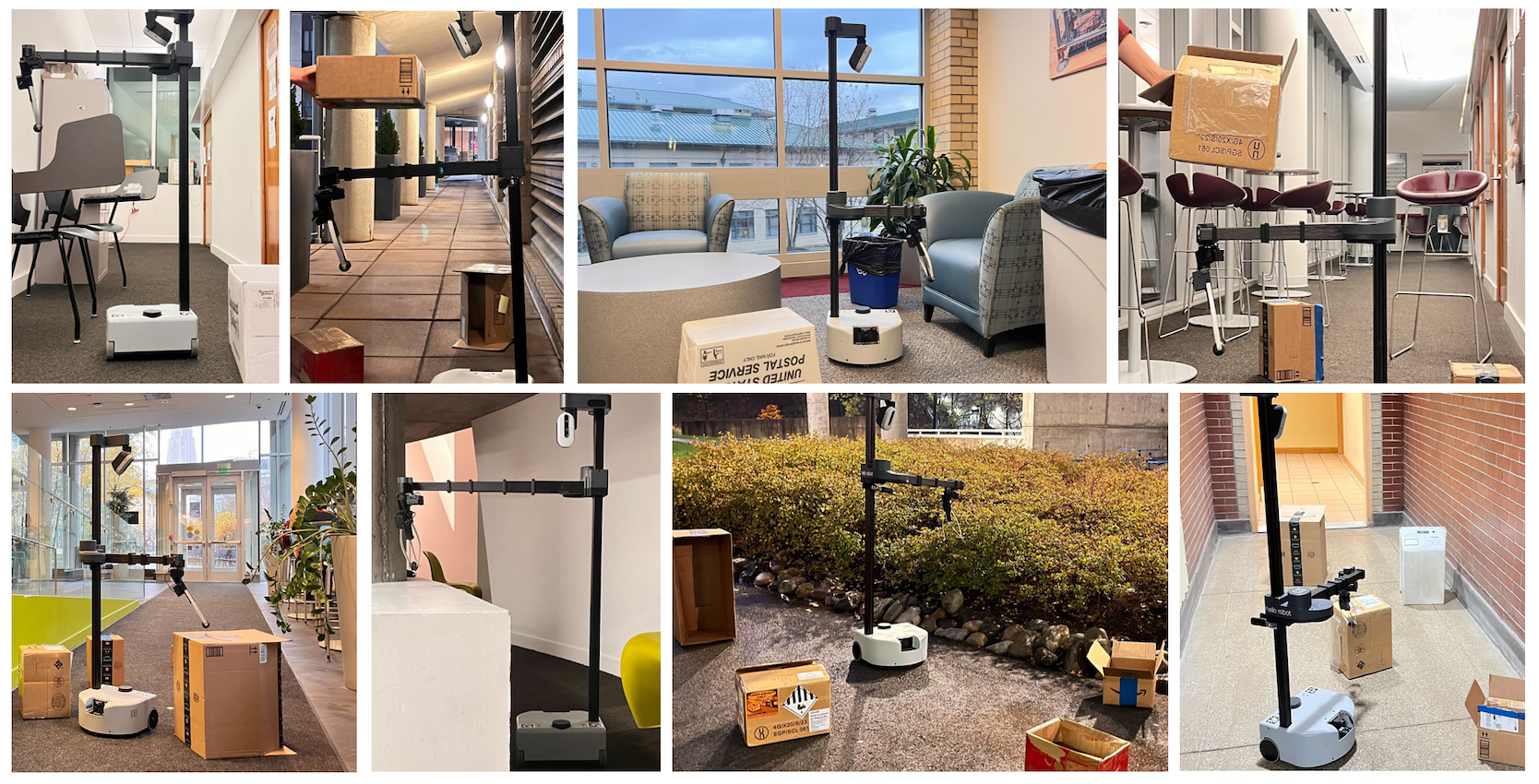}
    \vspace{-1em}
    \captionof{figure}{\small \textbf{Learning to SPIN:} Our robot learns to simultaneously perceive, manipulate, and navigate cluttered unstructured environments in a whole-body fashion. The robot has an actuated camera with a limited field of view that it must control to get information about its environment. The motion and perception problem are tightly coupled since what the robot knows about the environment influences how it can move and vice versa. We show results in a large variety of scenarios both indoors and outdoors with different obstacles like boxes and furniture. Our robot can pick up different objects like cups, and utensils. Video demos at \texttt{\href{https://spin-robot.github.io}{https://spin-robot.github.io}}
    }
    \label{fig:teaser}
\end{center}%
}]

\begin{abstract}
\vspace{-1em}
While there has been remarkable progress recently in the fields of manipulation and locomotion, mobile manipulation remains a long-standing challenge. Compared to locomotion or static manipulation, a mobile system must make a diverse range of long-horizon tasks feasible in unstructured and dynamic environments. While the applications are broad and interesting, there are a plethora of challenges in developing these systems such as coordination between the base and arm, reliance on onboard perception for perceiving and interacting with the environment, and most importantly, simultaneously integrating all these parts together. Prior works approach the problem using disentangled modular skills for mobility and manipulation that are trivially tied together. This causes several limitations such as compounding errors, delays in decision-making, and no whole-body coordination. In this work, we present a reactive mobile manipulation framework that uses an active visual system to consciously perceive and react to its environment. Similar to how humans leverage whole-body and hand-eye coordination, we develop a mobile manipulator that exploits its ability to move and see, more specifically -- to move in order to see and to see in order to move.  This allows it to not only move around and interact with its environment but also, choose ``when" to perceive ``what" using an active visual system.  We observe that such an agent learns to navigate around complex cluttered scenarios while displaying agile whole-body coordination using only ego-vision without needing to create environment maps. Videos are available at \texttt{\href{https://spin-robot.github.io}{https://spin-robot.github.io}}
\vspace{-2em}
\end{abstract}

\section{Introduction}
\begin{figure}[t]
    \centering
    \includegraphics[width=.96\linewidth]{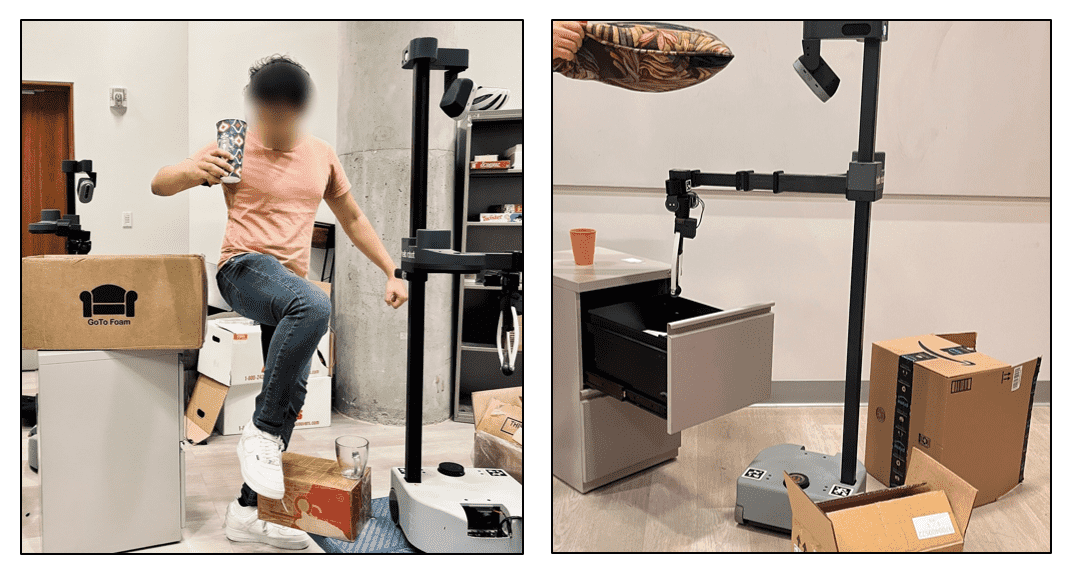}
    \vspace{-0.08in}
    \caption{\small Human and robot illustration of whole-body navigation through the clutter.}
    \label{fig:motivation}
    \vspace{-0.08in}
\end{figure}

Consider the example shown in Figure~\ref{fig:motivation}. A person is trying to carry a coffee cup through clutter. This not only requires navigational planning from start to goal but planning of the whole body to avoid obstacles along the way. Furthermore, due to ego-centric vision, the person needs to actively look around to gather the presence of obstacles. This general form of mobile manipulation task necessitates a coupled understanding of whole-body control with active perception. This capability is one of the fundamental and frequently encountered tasks in embodied cognition.

The dominant paradigm to tackle this problem is through classical planning-based control which requires apriori knowledge about the precise location of all the obstacles along with a detailed map of the environment. In most real-world scenarios, this assumption is impractical due to computational reasons, but more importantly, because environments are dynamic and objects keep moving around in general. Furthermore, relying on precise measurement of scenes for control does not allow agents to reactively improvise to changes in their environment. Practically, even when the complete environment map is known apriori, joint planning for a system with high degrees of freedom, say a mobile base with an arm, is often intractable and too expensive to be deployed in real-time. 

Humans, on the other hand, do not rely on precise known estimates of object locations and instead use ego-centric vision to navigate around obstacles in real-time. In an unfamiliar environment, where to look is informed by where they want to move (called `active perception'), and how they move in return determines what all they can see immediately afterward. 
This integrated mobility and perception allows us to see, adapt, and react to maneuver through unseen heavily cluttered environments.

This paper presents \textit{\textbf{SPIN}}, an end-to-end approach to \textit{\textbf{S}}imultaneous \textit{\textbf{P}}erception, \textit{\textbf{I}}nteraction, and \textit{\textbf{N}}avigation. We train a single model that not only outputs low-level controls for the robot body and arm but also predicts where should the robot's ego-centric camera look at each time step while moving its whole body by avoiding obstacles.
We train our approach via reinforcement learning (RL), and to get around the computational bottleneck of rendering depth images, we use a teacher-student training framework where robot behavior is first learned using RL with access to visible object scandots and then distilled into a policy that operates from ego-depth using supervised learning. 
We evaluate across 6 benchmarks in simulation ranging from easy, medium, and hard difficulty, and two real-world environments with a similar level of clutter as the hard environments in simulation and also add dynamic, adversarial obstacles. We find that our method outperforms classical methods and baselines which do not use active vision. We also observed emergent behaviors, including dynamic obstacle avoidance which the robot did not see during training time.  

Our approach presents a radical hypothesis that the traditionally non-reactive planning approach to whole-body control can indeed be cast into a reactive model -- i.e. --  single end-to-end policy trained by RL. Despite a big departure from optimal control literature, this hypothesis is not as surprising since agile whole-body coordination and fast obstacle avoidance in humans are developed into muscle memory over time. We now discuss our approach in detail.

\begin{figure*}[t]
    \centering
    \includegraphics[width=1.\textwidth]{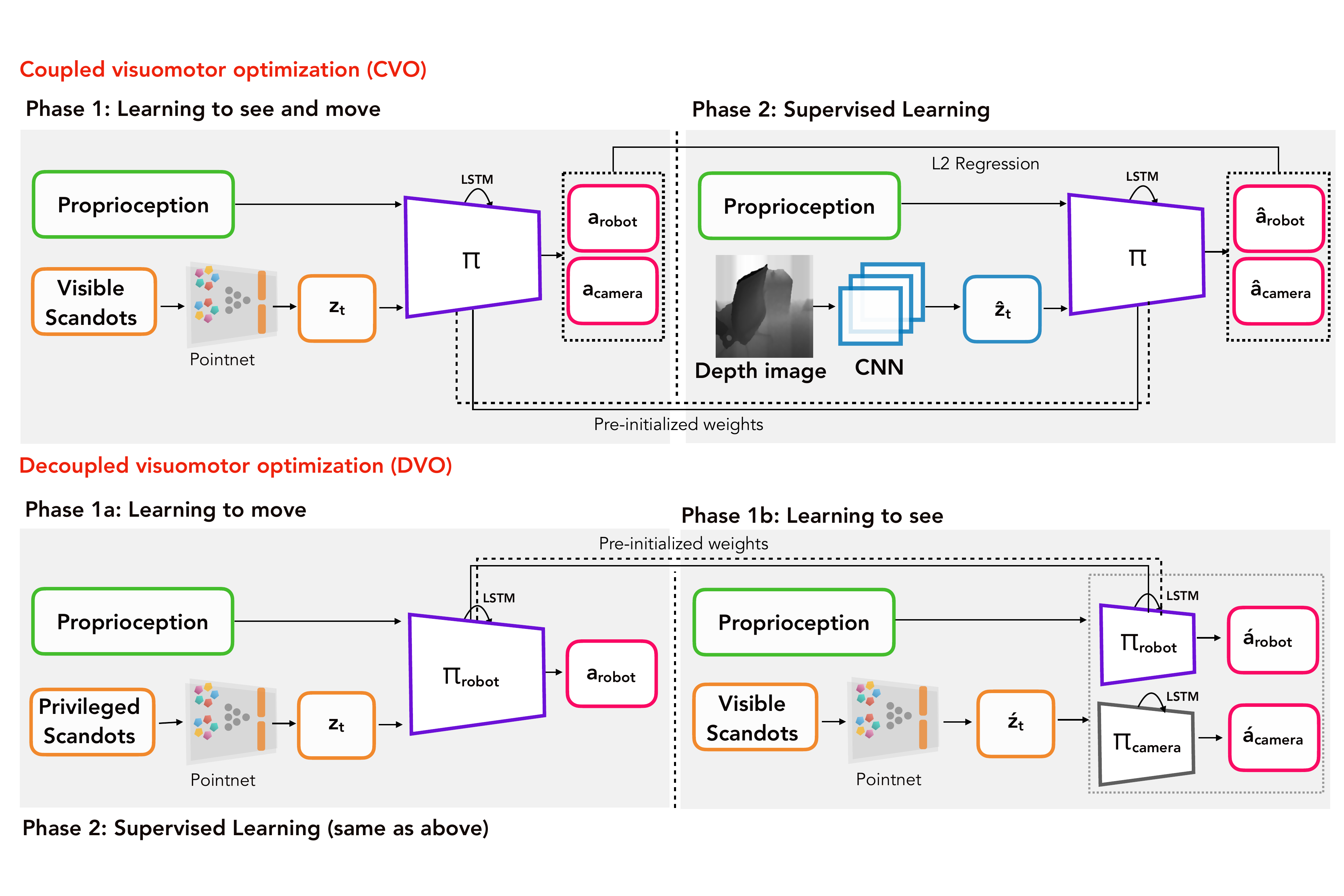}
    \vspace{-3em}
    \caption{We learn a policy that uses ego-vision to simultaneously perceive, interact, and navigate in cluttered environments. We propose two methods: (1) Coupled Visuomotor Optimization (CVO) learns robot and camera actions at the same time. We train an RL policy to predict these. We only provide scandots if they are visible in the agent's field-of-view allowing the agent to learn to move its camera and aggregate information about its environment. This is followed by a phase-2 supervised training where this behavior is distilled into a student network that operates with ego-centric depth images (2) Decoupled Visuomotor Optimization (DVO) decouples the action and perception learning into two parts: first the agent learns to navigate across clutter assuming access to all obstacles. In phase 1b, the robot learns to move its camera to estimate the relevant information. This is followed by supervised learning same as above.}
    \vspace{-1em}
    \label{fig:method}
\end{figure*}

\section{Method}
\label{sec:method}
We want our mobile manipulator (Fig.~\ref{fig:hello-robot}) to navigate and manipulate objects while avoiding obstacles in cluttered environments. It shares anatomical similarities with a human, bringing with it many of the same challenges. First, it has a limb in the form of an arm that can be raised and lowered, so the robot must constantly move the arm to avoid any obstacles. Second, it has an actuated camera with a very limited field of view ($87^\circ$ horizontal, $58^\circ$ vertical), so it needs to constantly look around to simultaneously plan ahead and look out for unexpected obstacles. Imagine yourself walking through a cluttered cabinet, there are too many obstacles around to keep track of, and you can't see all of them at once, so you must keep looking all around your body to plan a path through the clutter but also make sure you don't hit anything you missed along the way. Unlike regular walking where our eyes mostly point straight ahead and the path is clear, here you must \emph{actively} choose what to perceive for simultaneously planning ahead and also doing reactive fixes to your planned path. Since all the obstacles cannot be perceived at a single glance, you must have spatial awareness and know where the obstacle you saw some time ago is right now in relation to your body. Note that this entire process is very different from the classical approach, where perception, planning, and obstacle avoidance are separate processes executed separately and in sequence. Further, it is assumed that the output of each is perfect, whereas this is rarely the case in practice in our unstructured world.

To deal with this challenging, entangled problem setup, we take a data-driven approach. We train our robot to navigate inside procedurally generated clutter in simulation using RL. The robot is only allowed to perceive the part of its environment that is visible to the camera and learns to coordinate its arm, base, and camera motion such that it can plan ahead and reactively adjust to obstacles. 

In practice, since training with RL requires many samples and rendering depth is inefficient (see supp. Section \ref{sec:depth_vs_scandots}), we divide training into two phases. In the first one, we learn mobile manipulation behaviors via RL using a cheap-to-compute variant of depth and in phase 2 we train a CNN for perception from depth images as illustrated in Figure \ref{fig:method}.

\begin{figure*}[t!]
    \centering
    \includegraphics[width=1.0\textwidth]{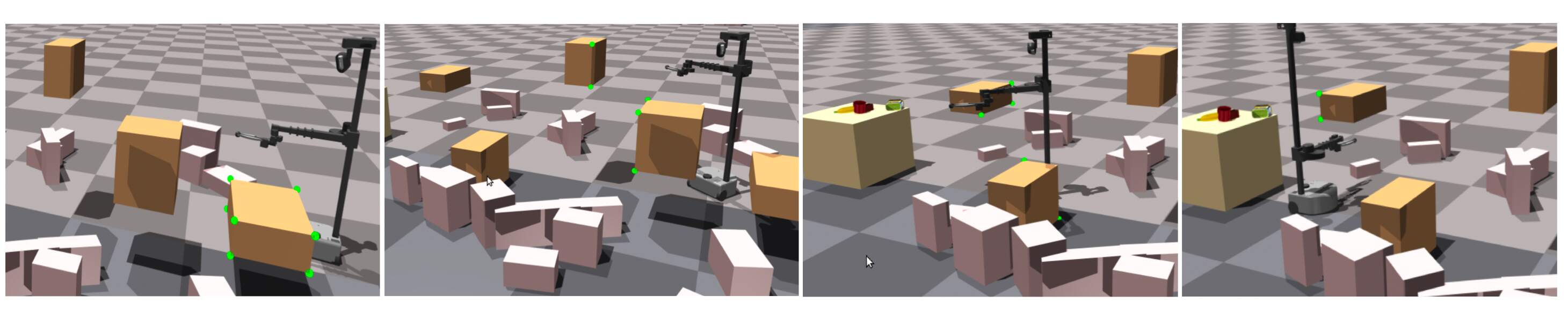}
    \caption{We illustrate one scenario of the simulation benchmark here with many obstacles in a narrow passage. The agent learns to develop whole-body coordination such as the robot's arm movement in the last two frames, to reactively adapt and navigate through such cluttered scenes by actively moving around the camera and aggregating information for efficient navigation without collisions.}
    \label{fig:sim_benchmark}
\end{figure*}

\subsection{Phase 1 - Learning Simultaneous Perception, Interaction and Navigation}
\label{sec:phase1}

In this stage, we use RL to learn to control all the joints of the robot to navigate clutter and pick target objects. Since rendering depth images directly from the robot camera is expensive, we must instead use an ersatz version that contains the same information and is cheap to compute. We do so using \textit{scandots} $\mathbf{s}_t$ which are the xyz coordinates of the bounding box of each obstacle. To specify which object to pick, we give the initial location of the object (before it is touched by the robot) $\mathbf{o}_i$. In lieu of the object image, we give the current location of the object $\mathbf{o}_t$. Here, scandots $\mathbf{s}_t$ and object location $\mathbf{o}_t$ are privileged information which must later be estimated from depth images. Given this information, we train two separate LSTM policies $\pi_\textrm{nav}$ and $\pi_\textrm{pick}$. At test time, the nav policy is activated to reach a target location and we switch to the pick one once the robot gets close to the object.
\vspace{-1.0em}
\subsubsection{Pick Policy} This accesses proprioception $\mathbf{x}_t$ consisting of robot joint angles and velocities $\mathbf{q}_t, \dot{\mathbf{q}}_t$, base linear and angular velocity $v_t, \omega_t$. For perception, it gets the object's initial and current location $\mathbf{o}_i$, $\mathbf{o}_t$ and predicts robot and camera actions. 
\begin{equation}
    [\mathbf{a}_\textrm{robot}, \mathbf{a}_\textrm{cam}] = \pi_\textrm{pick}(\mathbf{x}_t, F(\mathbf{o}_t, \mathbf{x}_t), \mathbf{o}_i)
\end{equation}
where $F$ is a masking function that masks object position $\mathbf{o}_t$ if it is not in the field of view of the camera. This is required since object position can only be estimated from depth in phase 2 if it is visible. 
\subsubsection{Navigation Policy} 
Training this policy requires a complex joint visuomotor optimization since robot motion is dependent on its knowledge of the environment which in turn depends on how the robot moves. We present two approaches to tackle this problem.

\textbf{Coupled Visuomotor Optimization (CVO)} Here, we set up a partially observable environment for the robot and let the RL algorithm do the joint optimization using large-scale data. In particular, the policy gets proprioception $\mathbf{x}_t$ and only visible scandots $\mathbf{\tilde{s}}_t = F(\mathbf{s}_t, \mathbf{x}_t)$ as observation and has to predict both the camera and the robot actions. Since the scandots are permutation invariant, we pass them through a trainable point-net architecture $P$ to obtain compressed latent $\mathbf{z}_t = P(\mathbf{\tilde{s}}_t)$ that we pass to the policy
\begin{equation}
    [\mathbf{a}_\textrm{robot}, \mathbf{a}_\textrm{cam}] = \pi_\textrm{nav}(\mathbf{x}_t, \mathbf{z}_t)
\end{equation}
This presents a tough optimization landscape because the observations at each step are strongly dependent on $\mathbf{a}_\textrm{cam}$. For instance, if the camera swivels around the observations at the next timestep may look completely different. Indeed, we observe that this requires billions of samples inside a GPU-accelerated simulator to optimize which may not always be feasible in practice.

\textbf{Decoupled Visuomotor Optimization (DVO)} To ease the optimization process, we learn the robot and camera actions separately. First, we learn how to move by giving the robot access to all available scandots $\mathbf{z}_t = P(\mathbf{s}_t)$ in a local vicinity. Since the robot sees everything, the camera motion is irrelevant and we just predict the robot's motion
\begin{equation}
    \mathbf{a}_\textrm{robot} = \pi_\textrm{nav}^\textrm
    {1a}(\mathbf{x}_t, \mathbf{z}_t, \mathbf{g}_t)
\end{equation}
where $\mathbf{g}_t$ is the goal with respect to the base. Using this policy as supervision, we train another policy to predict both camera and robot motions with access to only visible scandots $\mathbf{\hat{z}}_t = P(F(\mathbf{s}_t, \mathbf{x}_t))$. This policy is trained via RL to predict the robot actions from phase 1 policy $\mathbf{a}_\textrm{robot}$. This optimization forces the student policy to learn camera behaviors that capture information about the environment that is needed to move optimally. We initialize $\pi^\textrm{1b}_\textrm{nav}$ from the weights of $\pi^\textrm{1a}_\textrm{nav}$
\begin{align}
    \min_{\pi^\textrm{1b}_\textrm{nav}}\quad & \|\hat{\mathbf{a}}_\textrm{robot} - \mathbf{a}_\textrm{robot}\| \notag \\ 
    \text{s.t.} \quad & [\mathbf{\hat{a}}_\textrm{robot}, \mathbf{\hat{a}}_\textrm{cam}] = \pi^\textrm{1b}_\textrm{nav}(\mathbf{x}_t, \mathbf{\hat{z}}_t, \mathbf{g}_t)
\end{align}
This decoupled approach learns to move and see in separate phases which eases the optimization burden. In principle, the coupled optimization is better since it is possible that the 1a policy may learn to exploit privileged information in a way that the 1b policy cannot estimate it for any set of camera movements. However, in our setting, this did not turn out to be the case. 

We train using PPO \cite{schulman2017proximal} with backpropagation through time \cite{werbos1990backpropagation} in procedurally generated environments.

\textbf{Rewards:} For the navigation task, we use distance to goal reward $\|\mathbf{g}_t\|$ along with a forward progress reward $|\left(\mathbf{v}_t\right)_g|$  where $\left(\mathbf{v}_t\right)_g$ is velocity along the direction of the goal. 
\begin{equation}
    r_\textrm{nav} = 0.1 \cdot \|\mathbf{g}_t\| + 0.1 \cdot |\left(\mathbf{v}_t\right)_g|
\end{equation}
For the pick task, we provide an object reaching reward, i.e., the distance between the gripper and the object. This is followed by a lift reward if a successful grasp is detected (based on whether contact forces cross a threshold). 
\begin{equation}
    r_\textrm{pick} = 0.5 \cdot \|\mathbf{o}_t - \mathbf{p}_t\| + 0.5\cdot r_\textrm{lift}
\end{equation}
where 
\begin{equation}
    r_\textrm{lift} = \left(1 - \tanh\left(15 \cdot \left[(\mathbf{o}_t)_z\right]_+\right)\right)\mathbb{I}\left[\sum_i f_i > 10\right]
\end{equation}
where $[x]_+ = \max(x, 0)$ and $\mathbb{I}$ is the indicator function which forces the reward to be active only when object contact forces $f_i$ exceed 10N. 

\textbf{Training environments:} We procedurally generate long corridors with obstacles placed in between the robot and the goal. The initial joints and orientation of the robot are randomized. Near the edges of the corridors, we place randomized obstacles and walls to simulate distractors in the depth image. For the pick task, objects are spawned on tables of varying dimensions. We use five different objects - a banana, mug, can, foambrick, and a bottle. The episode is terminated if the robot reaches the goal or hits an obstacle/table.   
\subsection{Phase 2 - From Scandots to Depth} 
Scandots are not directly observable in the real world and must instead be estimated from the depth image. We train a convolution network $C$ to convert rendered depth images $\mathbf{d}_t$ to perception latents $\mathbf{\tilde{z}}_t$. This latent is passed to a student policy $\pi'$ to predict the actions $[\mathbf{\tilde{a}}_\textrm{robot}, \mathbf{\tilde{a}}_\textrm{cam}]$. This is supervised using L2 loss from the phase 1 actions. The weights for $\pi'$ are initialized using $\pi$. We train this policy using DAgger \cite{ross2011reduction}. For the navigation policy, we optimize
\begin{equation}
    \min_{C_\textrm{nav}, \pi'_\textrm{nav}} \left\| \pi'_\textrm{nav}(C_\textrm{nav}(\mathbf{d}_t), \mathbf{x}_t, \mathbf{g}_t) - \pi_\textrm{nav}(\mathbf{z}_t, \mathbf{x}_t, \mathbf{g}_t)\right\|
\end{equation}
Note that the teacher policy $\pi_\textrm{nav}$ can be trained using either the coupled or decoupled approach. Similarly, for the pick policy, we estimate current object position $\mathbf{o}_t$ from depth
\begin{equation}
    \min_{C_\textrm{pick}, \pi'_\textrm{pick}} \left\| \pi'_\textrm{pick}(C_\textrm{pick}(\mathbf{d}_t), \mathbf{x}_t, \mathbf{o}_i) - \pi_\textrm{pick}(\mathbf{z}_t, \mathbf{x}_t, \mathbf{o}_t, \mathbf{o}_i)\right\|
\end{equation}

\section{Experimental Setup}
\label{sec:exp_setup}

We use the Hello Robot Stretch \cite{hello} for all our experiments (Fig.~\ref{fig:hello-robot}). The robot has 10 actuated joints which include 2 degrees of freedom for the camera, 2 for base rotation and translation, 2 for the arm, 1 for the gripper fingers, and 3 for the dexterous wrist. An Intel D435i depth camera is mounted on the top of the robot head which is actuated using two motors. The learned policy operates at 10Hz and we do velocity control for the robot base and position control for all the other joints. Velocity control for the robot base allows us to perform simultaneous robot translation and rotation for more agile behavior. We train using IsaacGymEnvs \cite{makoviychuk2021isaac} using 8192 environments which takes 6 hours of training for phase 1 and 10 hours of training time for phase 2 on a RTX 3090. We compare against the following baselines: 

\begin{figure}[t]
    \centering
    \includegraphics[width=\linewidth]{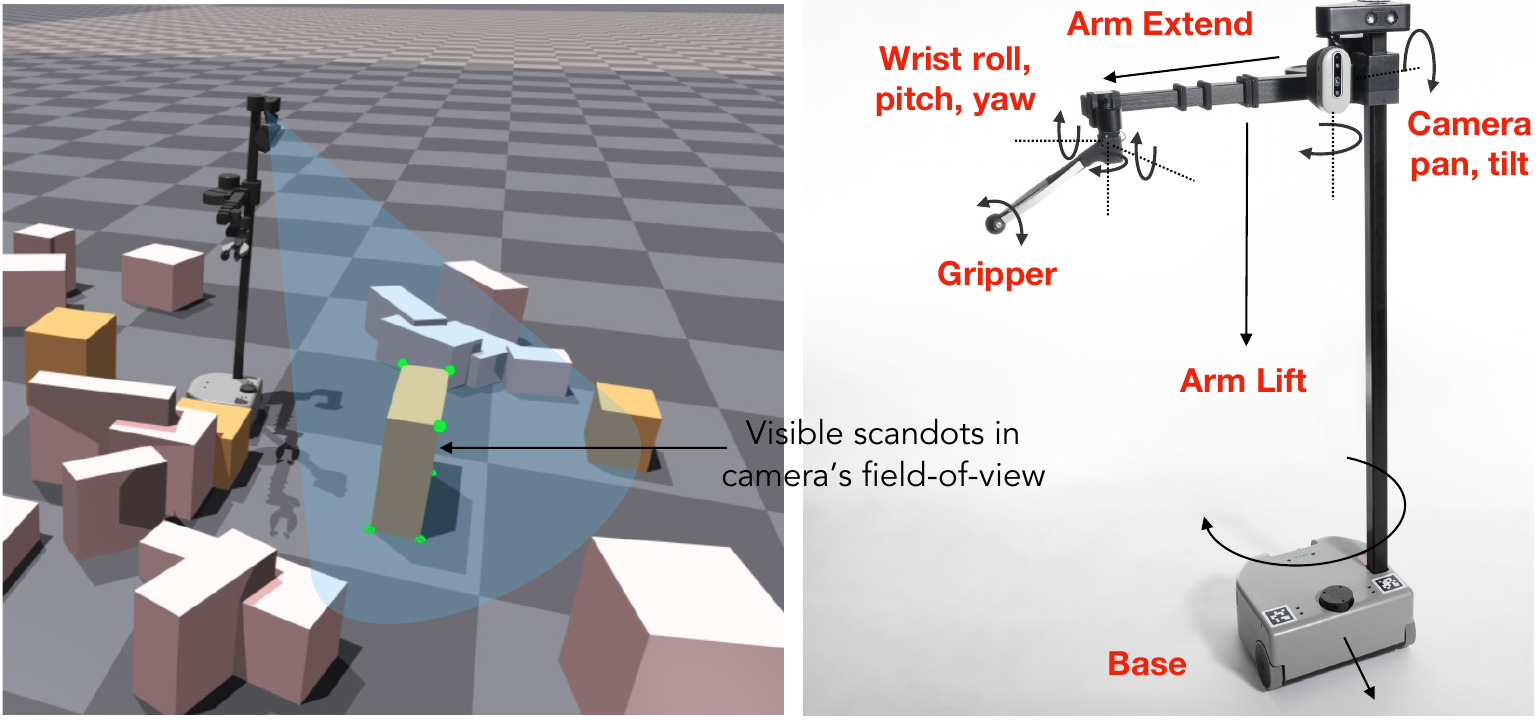}
    \caption{(Left) We compute visible scandots by projecting them to the camera frame and checking if they lie within the image plane (Right) of the stretch RE1 robot that we use experiments. It has two DoFs in the base, one each for arm lift and extension, two for the camera, three for the wrist, and one for the gripper.}
    \label{fig:hello-robot}
    \vspace{-1em}
\end{figure}

\begin{itemize}
    \item \textbf{FixCam:} The camera joints are frozen and the camera is forced to look forward. This baseline shows whether active vision is useful for the mobile manipulation problem and a fixed viewpoint is not enough. 
    \item \textbf{Mapping:} Instead of using a moving depth camera to get a series of frames this baseline assumes exteroception is provided in the form of a map. We simulate exteroceptive noise as in \cite{miki2022learning, agarwal2023legged}. 
    \item \textbf{Classical:} This uses a classical stack to control the base motion. 
    We first teleoperate the robot for 3-5 minutes to construct a map using the onboard 2D RPLidar using gmapping. Next, move\_base is used to plan a path through the environment. Finally, we move the robot to the start, use a Monte Carlo method \cite{1389727} to localize, and then execute the plan. Note that this baseline gets an easier version of the problem since it assumes that the map is known in advance and does not consider arm motion due to the 2D Lidar. This is used to test whether reactive navigation is superior to planning.  
    \item \textbf{NoPointNet:} Instead of passing object scandots through a permutation-invariant PointNet architecture, we concatenate them and use an MLP to estimate a latent.  
\end{itemize}

\section{Results and Analysis}

We evaluate our approach both in simulation as well as real-world. Since doing a lot of in-the-wild real-world experiments is more time-consuming and cumbersome due to various practical reasons, we thoroughly evaluate our approach on 6 simulation benchmarks with multiple scenarios. We explain each of these benchmarks in detail in Section \ref{sec:sim_results}.

While simulation benchmarks are useful for fair comparison with baselines as well as reproducibility, real-world experimenting is essential for determining the efficacy of our system in truly unstructured and dynamic environments. For this, we test our system on various real-world environments as shown in Figure~\ref{fig:teaser} and benchmark its performance on 2 real-world setups as described in Section~\ref{sec:real_results}. 

\begin{figure*}[t]
    \centering
    \begin{subfigure}[b]{1.0\linewidth}
        \centering
        \includegraphics[width=1.\linewidth]{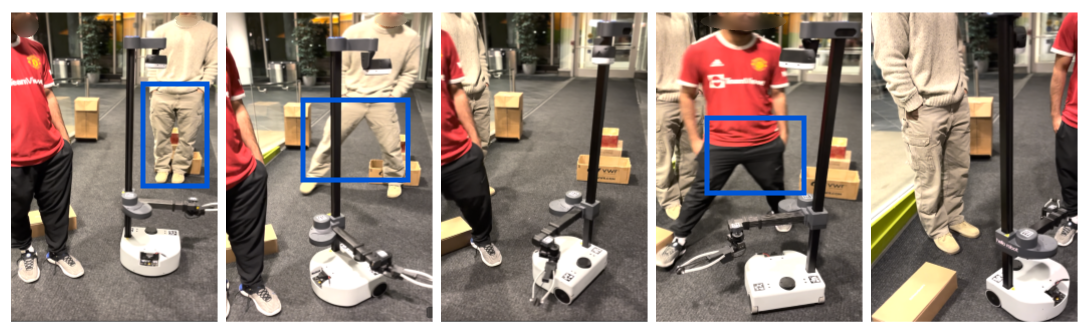}
        \caption{While our simulation lacks dynamic obstacles, the robot can still evade them because the policy continuously adjusts its plan.}
        \label{fig:sub1}
    \end{subfigure}
    \hfill
    \begin{subfigure}[b]{1.0\linewidth}
        \centering
        \includegraphics[width=0.99\linewidth]{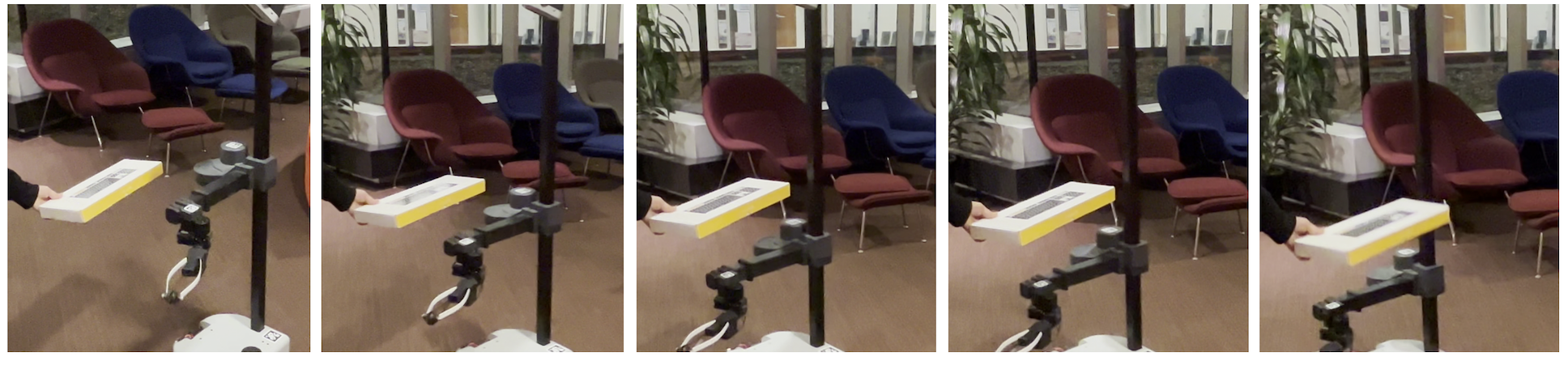}
        \caption{If there is an overhanging obstacle, the robot lowers its arm to avoid it instead of turning around, thereby displaying agile whole-body coordination.}
        \label{fig:sub2}
    \end{subfigure}
    \hfill
    \begin{subfigure}[b]{1.0\linewidth}
        \centering
        \includegraphics[width=1.\linewidth]{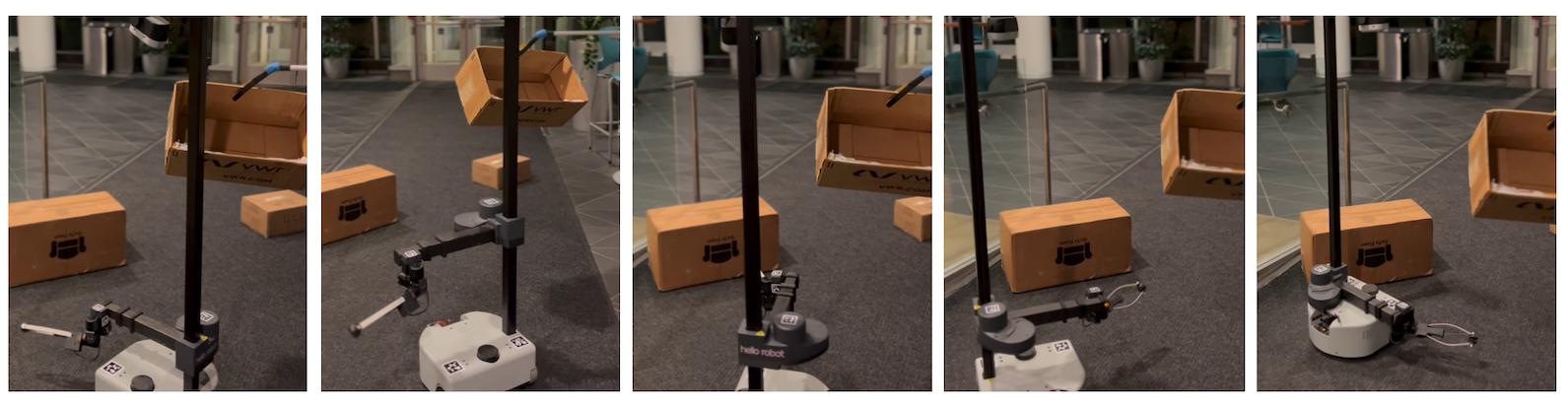}
        \caption{The robot adapts its plan on the fly to obstacles by turning around once the camera sees it.}
        \label{fig:sub3}
    \end{subfigure}
    \caption{Types of emergent behavior exhibited by \textit{SPIN} (a) dynamic obstacle avoidance (b) whole-body movement (c) adaptive rerouting.}
    \vspace{-0.5em}
    \label{fig:emergent}
\end{figure*}

Through simulation experiments, we aim to answer the following questions: (1) For a mobile agent, is active perception with an actuated camera necessary, or is a fixed viewpoint enough? (2) Can an active visual agent outperform a classical agent that relies on pre-built maps? What are the limitations of the latter? (3) What are some practical architectural design choices for optimizing mobility and perception together? We empirically answer each of these questions in Section \ref{sec:sim_results}.
Our real-world experiments primarily focus on comparing the capabilities of our reactive, learned system to a classical mapping and then planning approach. For our method, we observe interesting scenarios demonstrating emergent behaviors during real-world experiments detailed in Section \ref{sec:emergent} and \ref{sec:real_results}. 

\begin{table*}[h!]
\centering
\resizebox{\linewidth}{!}{
\begin{tabular}{p{1.5cm}cccccc|c|cccccc}
\toprule
& \multicolumn{6}{c|}{Reach} & Pick & \multicolumn{6}{c}{Place} \\
\cmidrule(r){2-7} \cmidrule{9-14}
& \multicolumn{3}{c}{Scenario 1} & \multicolumn{3}{c|}{Scenario 2} & & \multicolumn{3}{c}{Scenario 1} & \multicolumn{3}{c}{Scenario 2} \\
& Easy & Medium & Hard & Easy & Medium & Hard & & Easy & Medium & Hard & Easy & Medium & Hard \\
\midrule
\multicolumn{6}{l}{\emph{Success Rate}:} & & & & & & & \\
FixCam & 1.00 & 0.53 & 0.20 & 1.00 & 0.50 & 0.26 & 0.86 & 1.00 & 0.53 & 0.16 & 0.97 & 0.50 & 0.20 \\
NoPointNet & 1.00 & 0.87 & 0.57 & 1.00 & 0.77 & 0.63 & 0.93 & 1.00 & 0.83 & 0.57 & 1.00 & 0.77 & 0.60 \\
Mapping & 1.00 & 1.00 & 1.00 & 0.86 & 1.00 & 0.97 & 0.97 & 1.00 & 1.00 & 1.00 & 1.00 & 0.90 & 0.97 \\
\midrule
\textbf{\textit{SPIN}(DVO)} & 1.00 & 1.00 & 0.96 & 1.00 & 1.00 & 0.90 & 0.97 & 1.00 & 1.00 & 0.90 & 1.00 & 0.90 & 0.90 \\
\textbf{\textit{SPIN}(CVO)} & 1.00 & 0.97 & 0.93 & 1.00 & 1.00 & 0.93 & 0.97 & 1.00 & 0.97 & 0.90 & 1.00 & 0.97 & 0.93 \\
\midrule
\multicolumn{6}{l}{\emph{Average Episode Duration (s)}:} & & & & & & & \\
FixCam & 6.86 & 23.48 & 38.94 & 6.54 & 27.24 & 42.36 & 11.00 & 7.25 & 17.06 & 41.07 & 8.02 & 20.24 & 45.98 \\
NoPointNet  & 6.30 & 14.87 & 33.25 & 7.22 & 15.04 & 34.09 & 9.25 & 4.88 & 15.22 & 32.00 & 7.89 & 18.49 & 37.42 \\
Mapping  & 6.20 & 14.02 & 26.24 & 6.55 & 12.28 & 28.05 & 4.98 & 6.77 & 9.85 & 22.29 & 4.86 & 12.62
& 26.12 \\
\midrule
\textbf{\textit{SPIN}(DVO)}  & 5.92 & 16.25 & 28.44 & 7.32 & 17.12 & 32.04 & 9.24 & 7.22 & 18.45 & 34.24 & 9.40 & 15.98 & 41.24 \\
\textbf{\textit{SPIN}(CVO)}  & 6.24 & 14.00 & 23.57 & 6.55 & 15.81 & 29.31 & 5.74 & 6.51 & 13.39 & 27.25 & 8.03 & 12.79 & 31.25 \\
\bottomrule
\end{tabular}
}
\caption{We evaluate the success rate on 10 random environments with an average of 3 fixed seeds across all difficulty scenarios based on the obstacle course. We report the success rate of each part of the task including reaching \textit{(Reach)}, picking \textit{(Pick)}, and placing (\textit{Place)} the target object in the desired location. The \textit{place} task requires the agent to bring back the object across the obstacles near its start location.}
\label{tab:mytable}

\vspace{-1em}
\end{table*}

\subsection{Emergent Behavior}
\label{sec:emergent}
Large-scale simulation pre-training allows our robot to learn emergent behaviors to avoid obstacles in cluttered scenarios, even in the presence of dynamic obstacles. We see several such behaviors during real-world experimentation which were neither planned nor specifically trained for in simulation but emerge as a result of a large diversity of procedural environments seen during training. We illustrate three such scenarios in Figure \ref{fig:emergent}. As highlighted in several frames, Figure \ref{fig:sub1} depicts robustness to adversarially placed dynamic obstacles that constantly block the path of the robot. It needs to continuously perceive its environment in multiple directions and quickly react to those changes. We observe that in cases when there is no feasible path for the robot to navigate through, it also learns to stop and look around in order to replan its path and avoid collisions. Similarly, in \ref{fig:sub2} we see that as soon as a floating obstacle is suddenly placed in front of the robot, it shows spatial awareness and whole-body coordination and lowers its arm to navigate through, instead of turning and replanning the entire base movement which would take more time. In Figure \ref{fig:sub3}, we see an adaptive rerouting mechanism where the agent changes its straight-line motion as soon as a person kicks in a box in front of it. These behaviors emerge in real-time and show the ability of our system to continuously perceive, adapt, and react to changes in its environment which is very hard for a classical planner.

\subsection{Real-world results}
\label{sec:real_results}
We test on two real-world scenes - an academic lab and an open study area with couches and a kitchenette next to it with both static as well as dynamic obstacles. Both these environments have unstructured clutter and humans as dynamic obstacles that make it challenging for the agent to navigate through these spaces. For each environment, we have 4 static obstacles and at most 1 dynamic obstacle thrown adversarially. We compare against a classical baseline that uses an A1 RPLidar with gmapping and move\_base for planning. We first teleoperate the robot for 3-5min to create a map. Note that this provides the added advantage that this baseline knows the entire map in advance.  Since the Lidar cannot see objects above the plane we only test on ground obstacles and ignore floating ones that require whole-body coordination. We run the planner to only plan the base motion. In Tab.~\ref{tab:static_dynamic_baseline} we compare the success rate and average number of collisions. An episode succeeds when the robot reaches within 15cm of the specified goal position. Overall, our method succeeds 20-40\% more than the classical baseline. This is because the classical method suffers from noise and is not able to recover from a noisy map, and gets stuck, whereas the learned policy learns to look at the obstacles again and again to improve its uncertainty estimates and constantly updates its knowledge of where obstacles are. This ability is even more apparent in the dynamic scenario (Table ~\ref{tab:static_dynamic_baseline}) where the classical has a near zero success rate while our method can succeed. It has the emergent ability to avoid a new obstacle in space, whereas the classical baseline relies on the pre-built map and fails entirely. Note that, we do not train our policy with dynamic obstacles in simulations, but this behavior comes out as a by-product of lots of diverse experience in simulation. We design the observation space such that everything is relative to the robot. This allows the agent to perceive the environment as moving within its local reference frame, allowing generalization to dynamic obstacles.

\subsection{Simulation results}
\label{sec:sim_results}

The simulation benchmarks have 6 scenes, 2 of each easy, medium, and hard environments. Easy environments have 0-1 obstacles within a 5m goal range. Medium environments have 2-3 obstacles within 5m and the hard ones have heavily cluttered scenes with 5 obstacles within 5m. In each of these cases, one scene (Scenario 1) comprises of a tight 1m wide long corridor which bounds the agent to not take shortcuts and reach the goal only by navigating through obstacles. The second (Scenario 2) is an L-shaped corridor with the goal at the end. The evaluation metrics are reported as an average of 10 episodes with random agent and obstacle initialization across 3 seeds.

We compare against various baselines to study the impact of our design decisions in Tab.~\ref{tab:mytable}. For each scenario, we report the success rate and average episode length across 10 rollouts. Our method achieves $\approx$ 33\% higher success rate than the NoPointNet baseline since permutation invariant scandots latent makes the optimization problem easier and also generalizes better at test time. Ours achieves $\approx$ 68\% higher success rate than the FixCam baseline with the camera pointing straight ahead. This is because in some cases the robot encounters obstacles in its peripheral vision and our policy can change the camera angle to avoid them. Active vision is necessary for the robot to move effectively through a cluttered environment. Our method is significantly better than the Mapping baseline because the systematic noise in the object locations makes it hard for the robot to avoid them, especially in cluttered environments, whereas our method can continuously estimate the position of obstacles while it is moving and adapt the motion online. Finally, we compare the decoupled (DVO) optimization against coupled (CVO) optimization variant of our method and find that they achieve similar performance. We hypothesize that the partial observability and joint optimization for camera and robot actions in CVO training allows the agent to quickly discover optimal shortcuts that are otherwise harder to distill from a privileged teacher policy.

\begin{table}
    \centering
    \resizebox{\linewidth}{!}{
    \begin{tabular}{lcc|cc}
        \toprule
        & \multicolumn{2}{c}{Static Obstacles} & \multicolumn{2}{c}{Dynamic Obstacles} \\
        \midrule
        \multicolumn{5}{c}{Scenario 1} \\
        \midrule
        & Ours & Classical & Ours & Classical \\
        Average Success & 0.8 & 0.6 & 0.6 & 0.0 \\ 
        Average \# Collisions & 1.0 & 0.4 & 1.6 & 1.2 \\
        \midrule
        \multicolumn{5}{c}{Scenario 2} \\
        \midrule
        & Ours & Classical & Ours & Classical \\
        Average Success & 0.8 & 0.4 & 0.6 & 0.2 \\ 
        Average \# Collisions & 0.8 & 0.6 & 1.6 & 1.0 \\
        \bottomrule
    \end{tabular}
    }
    \caption{We compare our method against a classical mapping and  planning baseline for navigation in cluttered scenes with both static as well as dynamic obstacles. The classical performs reasonably in static environments, it quickly breaks with dynamic obstacles like humans walking around, whereas our method shows more robust reactivity to such obstacles even without being trained with dynamic obstacles in simulation. We report the success rate of our method compared with the baseline. For the classical baseline, we teleoperate the robot for 2-3 min.}
    \vspace{-1em}
    \label{tab:static_dynamic_baseline}
\end{table}

\section{Related Works}

\label{sec:related_works}
\paragraph{Classical Approaches}
The problem of navigating robots around obstacles has been studied for decades. Classical methods solve the motion and perception problem separately. First these methods build a map of the environment using the robot's onboard sensors such as cameras, proprioception and Lidar or infrared \cite{cassandra1996acting, gutmann1998experimental}. Kalman-filter-based \cite{welch1995introduction} techniques are often used to track positions, but they can't represent multi-modal ambiguities or recover after tracking failure \cite{roumeliotis2000bayesian}. Grid-based methods solve this but suffer from high memory usage \cite{burgard1996estimating}. Modern SLAM approaches ORBSLAM3 \cite{mur2015orb}, OpenVSLAM \cite{sumikura2019openvslam} and RTAB \cite{labbe2019rtab} use variations of a method  that relies on particle filters \cite{THRUN200199} to hold a multi-modal belief of the robot's location in the map \cite{thrun2002particle, liu1998sequential}. SLAM is especially challenging in dynamic environments due to the confounding motion of other agents  \cite{yu2018ds, saputra2018visual, Fox1999markov, wolf2005mobile}.  Once a map is built, a path can be planned over it. Exact paths can be computed using graph search algorithms \cite{hart1968formal}, probabilistic methods which are faster but yield approximately optimal solutions \cite{lavalle2001rapidly} or potential-field based methods \cite{khatib1986potential}. All of these assume perfect perception and re-planning is usually expensive making them susceptible to noise and precluding reactive behavior. 

\vspace{-0.06in}
\paragraph{Learning-based navigation}
In recent years, learning has been used to improve the classical navigation stack. Modular approaches \cite{chaplot2020learning, chang2023goat, gupta2017cognitive, luo2022stubborn} still leverage SLAM-based methods to build a map but use learning or heuristic changes to get priors for the best possible route to a goal. End-to-end approaches forgo maps entirely and train a policy to go from images to robot commands to go to a goal location \cite{wijmans2019dd, chattopadhyay2021robustnav, chen2018learning}. We also take the end-to-end approach but unlike prior work where what the robot sees is fixed based on its position, in our case it must move its head and actively choose what it sees making optimization more challenging.

\vspace{-0.06in}
\paragraph{Mobile Manipulation}
A mobile base and arm together can complete useful in-the-wild manipulation but present a more challenging control problem. Imitation learning techniques focus on collecting large datasets in a variety of settings with a dexterous 6-dof arm and a wheeled mobile base using teleoperation \cite{wong2022error, pmlr-v162-du22b, brohan2022rt, ahn2022can, burgess2023enabling, haviland2022holistic}. Because of the high-dimensionality of mobile manipulation, there is also control methods that leverage synergies between both the base and the arm and plans together.   \cite{yokoyama2023adaptive,haviland2022holistic, hu2023causal, fu2023deep}. 

\section{Discussion and Limitations}
We present \textit{SPIN}, an approach to train robots that can simultaneously perceive, interact, and navigate cluttered environments using a data-driven approach. We show that our RL-based reactive approach is effective for active whole-body control-perception problem, traditionally addressed via non-reactive planning methods. With recent interest in humanoid and other mobile robots with actuated cameras, on neck for instance, \textit{SPIN} is a cost-effective agile whole-body control solution with limited sensing and compute.

Although our robot can perceive geometry and avoid obstacles using depth, it still operates on stereo-matched depth instead of raw RGB. This leads to scenarios where it can bump into glass obstacles or shiny surfaces. In the future, we would like to use RGB for perception.

\vspace{-0.65em}
\paragraph{Acknowledgements} We thank Jared Mejia and Mihir Prabhudesai for helping with stress-testing in real-world experiments. We are also grateful to Zackory Erickson and the Hello Robot team for their support with the robot hardware. This work was supported in part by grants including ONR N00014-22-1-2096, AFOSR FA9550-23-1-0747, and the Google research award to DP.

\label{sec:discussion}
 
{
    \small
    \bibliographystyle{ieeenat_fullname}
    \bibliography{main}
}

\clearpage
\setcounter{page}{1}
\maketitlesupplementary

\section{Qualitative Results}
\label{sec:qualitative_results}

We test our framework in several in-the-wild scenarios, some of which are illustrated in Figure \ref{fig:teaser}. Qualitative video results are available at \texttt{\href{https://spin-robot.github.io/}{https://spin-robot.github.io/}}

We see emergent behavior where the robot continuously avoids dynamic obstacles without seeing them during training. We also observe generalization heavily cluttered indoor to dim-lit outdoor environments. The agent also demonstrates reactive whole-body coordination where it moves its arm up or down to efficiently navigate across floating obstacles instead of re-routing and re-planning base movement, demonstrating 3D spatial awareness. 

\begin{figure*}
    \centering
    \subcaptionbox{\small{Pose I: Facing forward\label{fig:sub1}}}{\includegraphics[width=0.24\linewidth]{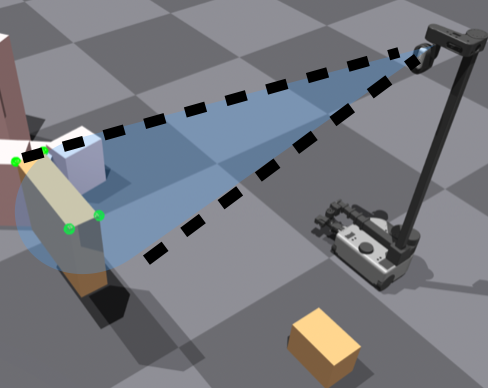}}\hfill
    \subcaptionbox{\small
{Pose II: Facing downward\label{fig:sub2}}}{\includegraphics[width=0.24\linewidth]{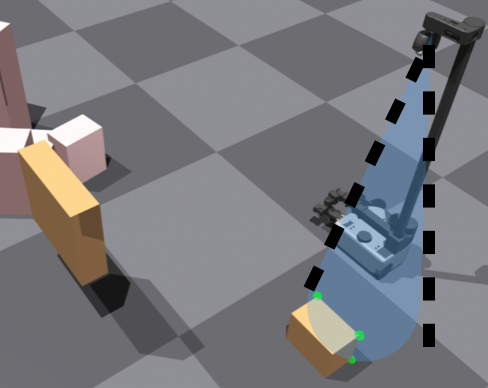}}\hfill
    \subcaptionbox{\small{Pose III: Facing downward and slightly forward\label{fig:sub3}}}{\includegraphics[width=0.24\linewidth]{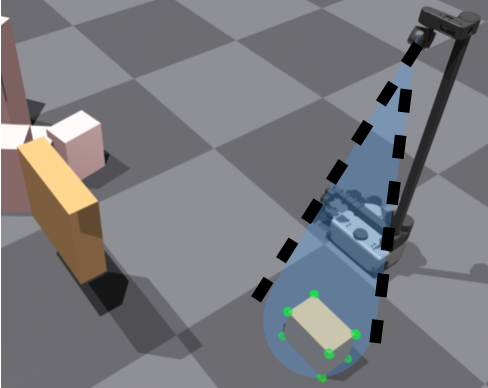}}\hfill
    \subcaptionbox{\small{Pose IV: Facing forward and slightly downward\label{fig:sub4}}}{\includegraphics[width=0.24\linewidth]{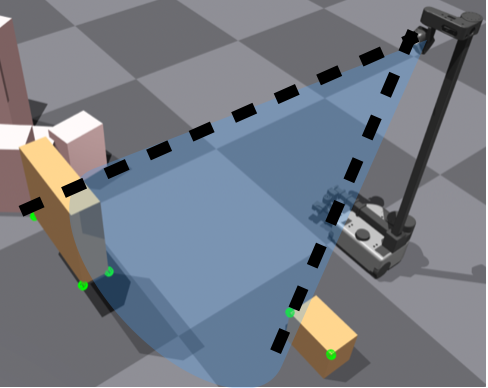}}
    \caption{Different camera poses for the FixCam baseline.}
    \label{fig:fixcam_baselines}
\end{figure*}

\section{Implementation Details}
\label{sec:implementation_details}

We make several design choices for the working of our framework. Firstly, the robot is only allowed to have local visibility in order to develop highly reactive and instant behaviors. At any time instant $t$, the agent can perceive it's environment within a range of 2m in all 4 direction -- front, back, left and right based on the camera's viewing direction and it's field of view. We also empirically observe that given larger viewing range, say \(>\) 5m which contains information of more than 4-5 nearest obstacles to the agent makes it a sub-global path planning problem which becomes harder to optimize, leading to degraded behavior and performance as reported in Table \ref{tab:local_visibility}.

\begin{table}[htbp]
    \centering
    \begin{tabular}{c|c|c}
        Visibility Range & Success Rate $\uparrow$ & Distance to Goal (m) $\downarrow$ \\
        \hline
        \( \leq 1 \)m & 0.96 & 0.28 \\
        \( \leq 2 \)m & 0.96 & 0.26 \\
        \( \leq 3 \)m & 0.93 & 0.63 \\
        \( \leq 5 \)m & 0.86 & 1.21 \\
    \end{tabular}
    \caption{We report the average success rate and average distance to goal for 10 episodes across 3 seeds each with different maximum visibility range for the agent at any time instant. As reported, with broader visibility, the agent shows more frequent stalling leading to higher average distance to goal.}
    \label{tab:local_visibility}
\end{table}

Secondly, contrary to standard teacher-student architectures which use privileged system information for training the teacher, we restrict the privileged information to only elements in the robot's field-of-view that can be retrieved from the ego-view in the same state. For this, we project the obstacle scandots onto the image plane and pass only those scandots to the robot observation which lie in the camera's field-of-view. Note that, if this were not the case, the robot does not learn camera movement in a relevant fashion and also becomes harder to distill into a depth-conditioned student policy through only ego-centric view. Similarly, for the 3-phase decoupled visuo-motor optimization, we induce information bottleneck with a low-dimensional latent space of size 16 for the scandots latent while training the teacher policy, which again helps it in attending to most relevant information at any time instant $t$ in order to make student policy distillation feasible. 

\paragraph{Observation Space.} The observation space for the robot comprises of joint positions $(q)$, joint velocities $(q_vel)$, end-effector position $p_{eef}$, goal position $(p_{goal})$ and depth latent $(\hat{z})$ containing visual information about the environment. Note that during phase 1 training in simulation, scandots $(z)$ are used as a proxy for faster depth rendering and later distilled into a egocentric depth-conditioned policy. During real-world deployment, $p_{eef}$ is obtained via forward kinematics and other proprioception information is obtained directly from the robot.

\paragraph{Action Space.}

The action space of the robot consists of the velocity for base rotation as well as translation and joint positions for all the other joints including arm, camera as well as gripper actions. The gripper action is a continuously varying scalar which can actuate the gripper to different extents, unlike a binary action indicating open or closed gripper.

\paragraph{Reward Scales.}
We use a distance and forward progress goal for reaching, a binary reward for grasping and a continuous shaped reward for lifting the object to a certain height above the table. We also add a small penalty for the joint velocities to the arm stretch and camera joints for a temporally smooth gait which permits easier sim2real transfer as well as more appropriate behaviors leading to less jitter and more consistent movements on the real hardware. 

The reward scales used for goal reaching, grasping and lift rewards are reported in Table \ref{tab:reward_scales}. Detailed formulations of the reward functions are described in Section \ref{sec:phase1}.

\begin{table}[htbp]
    \centering
    \begin{tabular}{c|c}
         & Reward Scale \\
        \hline
        Reach Reward & 0.1 \\
        Grasp Reward & 0.5 \\ 
        Lift Reward & 0.8 \\ 
        Joint Velocity Penalty & -0.03 \\  
    \end{tabular}
    \caption{We report the average success rate and average distance to goal for 10 episodes across 3 seeds each with different maximum visibility range for the agent at any time instant. As reported, with broader visibility, the agent shows more frequent stalling leading to higher average distance to goal.}
    \label{tab:reward_scales}
\end{table}

\paragraph{Network Architecture and Training Details.}

The actor and critic for teacher policy are LSTM with 256 hidden units, with input as prorpioception, goal and scandots latent. The scandots are compressed using a pointnet architecture for permutation invariance. The depth network for the student policy takes as input a low-resolution depth image of size $58 \times 87$ and comprises of 3-layer convolution backbone followed by 3 fully-connected layers. We use Adam Optimizer with an initial learning rate of $1e-3$, entropy coefficient of $5e-4$ and $\gamma$ as $0.99$.

\paragraph{Asynchronous DAgger Training. } Since depth rendering on simulators is a computational bottleneck, we implement an asynchronous version of DAgger algorithm which simultaneously collects data in a buffer and trains the student policy with batches sampled from the collected data using 2 parallel processes. This provides a $2.5 \times$ computational speedup over the non-parallelized version of the algorithm, allowing faster convergence of the student network. We also find that freezing the weights of the student actor pre-initialized from the teacher policy for first 1000 iterations helps as warm-up steps to the depth convolution backbone for stable training.

\paragraph{Post-processing for clean depth images.}

To mitigate the issues due to noisy depth, we post-process the depth obtained from the Intel RealSense Camera using a real-time fast hole-filling algorithm for depth images \cite{ku2018defense}. With the camera constantly in motion, there are additional artefacts with depth images. For this, we additionally use temporal filtering over the stream of depth images.

\begin{figure*}
    \centering
    \includegraphics[width=1.0\linewidth]{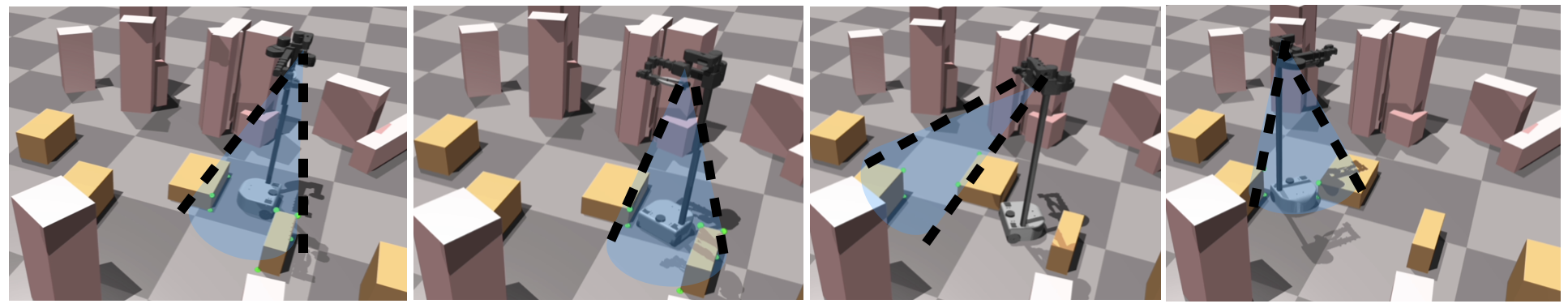}
    \caption{Camera movement analysis in a trajectory. The agent faces the camera downward when navigating through tightly cluttered vicinity as can be seen in the first, second and fourth frame, whereas the camera points more towards the front when there are no immediate obstacles in the direction of movement, as illustrated in the third frame.}
    \label{fig:cam_behavior}
\end{figure*}

\begin{figure*}
    \centering
    \includegraphics[width=1.0\linewidth]{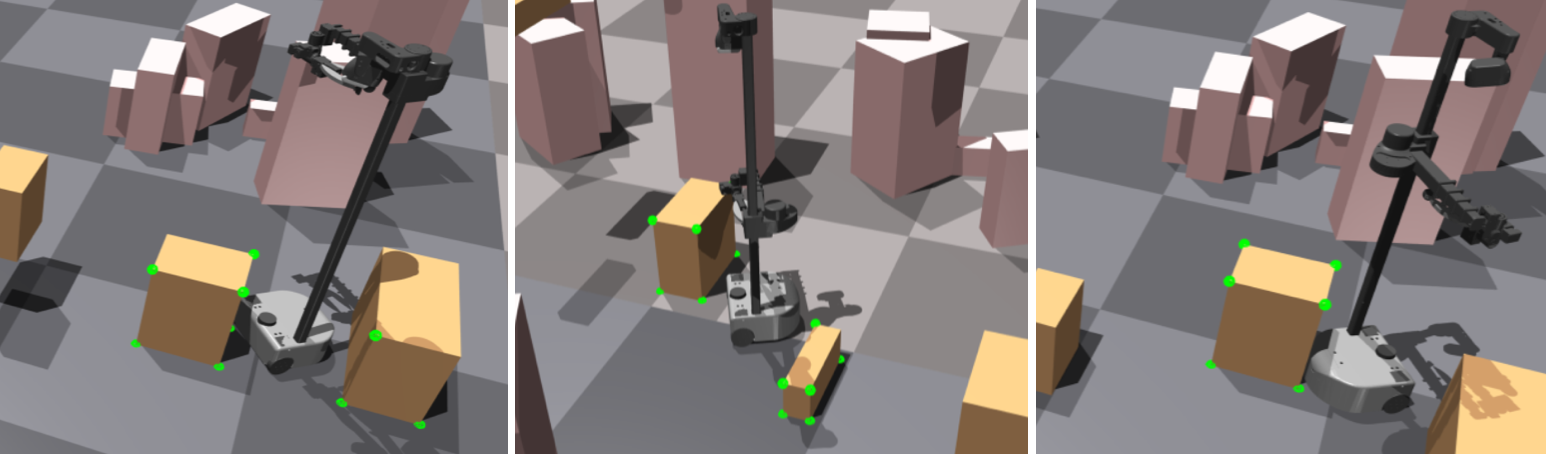}
    \caption{Scenarios where whole-body coordination is essential under heavy obstructions. In the above cases where the obstacles are tightly packed, it is not possible for the robot to navigate through them avoiding collisions without lifting the arm to an appropriate height.}
    \label{fig:whole_body_coord}
\end{figure*}

\paragraph{Object Detection for Pick Policy}

Once the robot reaches near the goal, we randomly select an object within its field of view in order to be grasped and fetched to a target location. For getting the target object location, we run YOLO \cite{Jocher_YOLOv5_by_Ultralytics_2020}, a real-time object detection model with an average inference speed of 20ms. We use the corresponding depth image to deproject the pixel point into a 3D-coordinate which is passed as the new goal position to the manipulation policy.

\section{Analysing camera and base motion}

\paragraph{Fixed Camera Baseline} We~run FixCam baseline with 4 camera poses (Figure \ref{fig:fixcam_baselines}) -- \textbf{I:} Front, \textbf{II:} Down, \textbf{III:} Down and slightly front, \textbf{IV:} Front and slightly down on easy (E), medium (M), hard (H) environments. \textbf{I}, \textbf{IV} with max fov have much lower success than SPIN, implying active vision is required in clutter. Pose (\textbf{IV} depicts the FixCam baseline referred in the paper.

\begin{table}
    \centering
    \begin{tabular}{c|ccc|ccc}
        \multirow{2}{*}{} & \multicolumn{3}{c|}{ Scenario 1} & \multicolumn{3}{c}{ Scenario 2} \\
        & E &  M & H &  E & M &  H \\ 
        \hline
        \textbf{I} & 0.93 & 0.40 & 0.20 & 0.97 & 0.40 & 0.20 \\
        \textbf{II} & 0.70 & 0.30 & 0.10 & 0.67 & 0.47 & 0.10 \\
        \textbf{III} & 0.86 & 0.33 & 0.10 & 0.77 & 0.30 & 0.10 \\
        \textbf{IV} & 1.00 & 0.53 & 0.20 & 1.00 & 0.50 & 0.26 \\
    \end{tabular}
    \caption{ Success rate for 4 FixCam poses in easy (E), medium (M), hard (H) envs.}
    \label{tab:fixcam}
\end{table}

\paragraph{Camera Movement and Camera Observations:} We show camera trajectory in Figure \ref{fig:cam_behavior}. When navigating through clutter (frames 1, 2, 4), it tilts downward to maximize fov near the base, but with no nearby obstacle (frame 3), it faces \textit{front}. Detailed movement of the camera can be seen on the website along with paired RGBD images for rollouts. RGB frames are only for analysis, the policy only observes depth images.

\paragraph{Necessity for Active Vision and Whole-body coordination}

\noindent \textit{Active Vision:} In principle, a multi-camera system should be equivalently adequate, however most views will contain insignificant information and require large models to process. With limited onboard compute on most robots and requirement for real-time reactivity ($<0.1s$), it becomes infeasible to deploy them with larger vision backbones.

\noindent \textit{Whole-body coordination (WBC):} Under heavy obstructions, the robot cannot move without collision if the base \& arm control are decoupled. Figure \ref{fig:whole_body_coord} shows such a scenario where a fixed arm and gripper close to base would fail without WBC, which also allows it to use the extra degree of freedom to find shorter and more efficient paths.

\section{Directly training from depth images.}
\label{sec:depth_vs_scandots}
\begin{figure}
\centering
\includegraphics[width=0.6\linewidth]{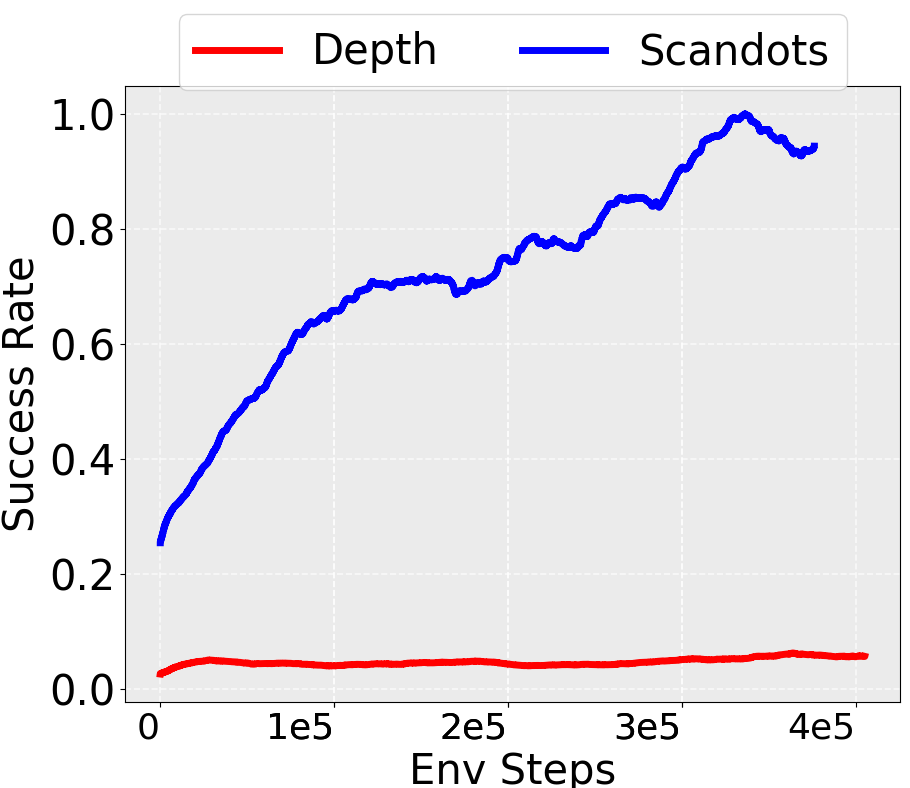}
\caption{Success rate for scandots (blue) vs depth (red). The depth-based policy attains close to 0 performance even after 400k env steps of training, whereas the policy trained with scandots increasingly improves over time.}
\label{fig:depth_vs_scandots}
\end{figure}

We compare training from depth (red) and our 2-phase method with scandots (blue) in a medium  difficulty environment as illustrated in Figure \ref{fig:depth_vs_scandots}. Depth policy has $<1\%$ success after 22h training, whereas total (\textit{phase1} + \textit{2}) wall-clock time for SPIN is 16h (6+10). The simulator gives $\approx50k$ fps for scandots (8192 envs) and $\approx820$ fps for depth (256 environments -- maximum parallel environments that can fit on a single GPU), causing $61\times$ slow-down bottleneck. This shows the necessity and efficiency of our proposed 2-phased coupled visuomotor optimization approach using scandots over naively training an RL policy from depth observations.

\section{Classical Navigation Baseline}
\label{sec:baseline_details}

\begin{figure*}
    \centering
    \includegraphics[scale=0.5]{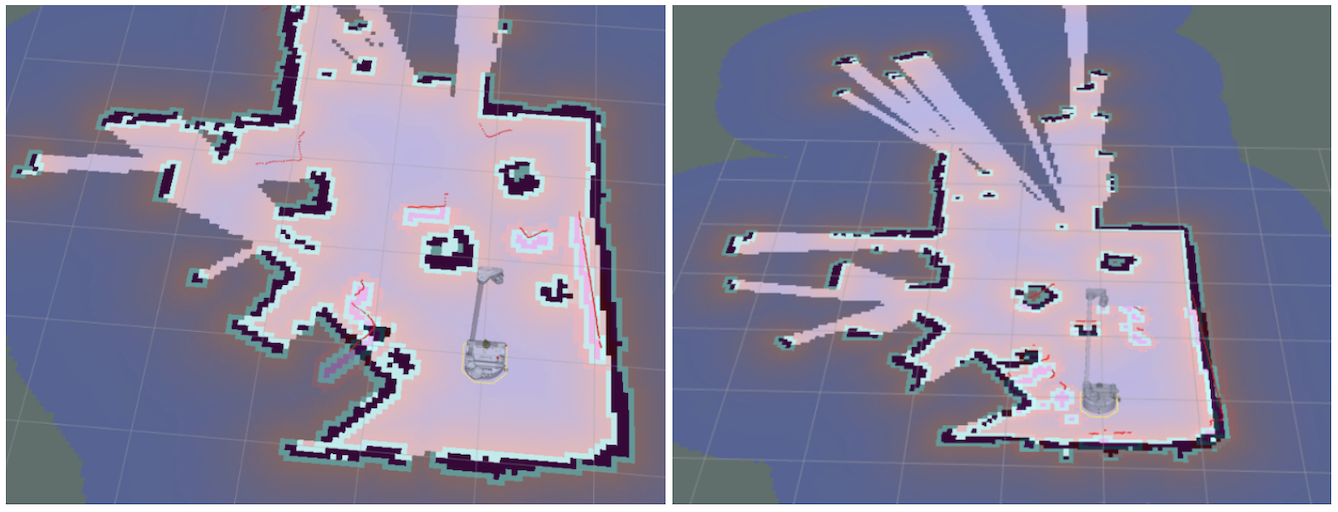}
    \caption{Visualizations of environment map built using 2D Lidar. The robot is localized as per its initial position and orientation.}
    \label{fig:map_viz}
\end{figure*}

As discussed in Section \ref{sec:exp_setup}, we compare our method with a classical map-based baseline which uses the 2D RPLidar to build an environment map and then creates a plan using Monte Carlo method. We observe that for $ >90\%$ of the cases, the robot is able to build a map and find a feasible path, however it is not able to execute the planned path for $>85\%$ cases. This issue arises due to noisy control or unexpected wheel motion due to terrain differences. Our method is able to overcome such failures due to a constant feedback and reactive improvisation through proprioception as well as depth, which allows it to deal with uncertainties without requiring a pre-built environment map. Moreover, due to localization inaccuracies, the baseline method is often unable to reach the intended goal if not initialized in the same orientation as was used before building the map. Contrary to that, we heavily randomize all degrees of freedom as well as the robot orientation at the beginning of every rollout during test time. Moreover, since the robot has a 2D Lidar installed on it, we do not test it in environments with floating obstacles which would require 3D understanding and whole-body coordination to navigate through clutter. We show some visualizations of the map built and plans created by the robot in Figure \ref{fig:map_viz}.

\end{document}